\documentclass[nohyperref]{article}

\usepackage{microtype}
\usepackage{graphicx}
\usepackage{booktabs} 

\usepackage{hyperref}

\usepackage[accepted]{icml2022}

\usepackage{amsmath}
\usepackage{amssymb}
\usepackage{mathtools}
\usepackage{amsthm}

\usepackage[capitalize,noabbrev]{cleveref}

\usepackage{xcolor}
\usepackage{amsfonts}
\usepackage{amsmath}
\usepackage{booktabs}
\usepackage{siunitx}
\usepackage{subcaption}
\usepackage{graphicx}
\usepackage{mwe}

\theoremstyle{plain}

\theoremstyle{definition}

\theoremstyle{remark}

\usepackage[textsize=tiny]{todonotes}

\icmltitlerunning{}

\begin{document}

\twocolumn[
\icmltitle{Time-Series Anomaly Detection \\with Implicit Neural Representation}



\icmlsetsymbol{equal}{*}

\begin{icmlauthorlist}
\icmlauthor{Kyeong-Joong Jeong}{equal,yyy}
\icmlauthor{Yong-Min Shin}{equal,yyy}

\end{icmlauthorlist}

\icmlaffiliation{yyy}{Computational Science and Engineering, Yonsei University, Seoul, South Korea}

\icmlcorrespondingauthor{Kyeong-Joong Jeong}{jeongkj@yonsei.ac.kr}
\icmlcorrespondingauthor{Yong-Min Shin}{jordan3414@yonsei.ac.kr}


\vskip 0.3in
]



\printAffiliationsAndNotice{\icmlEqualContribution} 

\begin{abstract}
Detecting anomalies in multivariate time-series data is essential in many real-world applications. Recently, various deep learning-based approaches have shown considerable improvements in time-series anomaly detection. However, existing methods still have several limitations, such as long training time due to their complex model designs or costly tuning procedures to find optimal hyperparameters (e.g., sliding window length) for a given dataset. In our paper, we propose a novel method called Implicit Neural Representation-based Anomaly Detection (INRAD). Specifically, we train a simple multi-layer perceptron that takes time as input and outputs corresponding values at that time. Then we utilize the representation error as an anomaly score for detecting anomalies. Experiments on five real-world datasets demonstrate that our proposed method outperforms other state-of-the-art methods in performance, training speed, and robustness.
\end{abstract}

\section{Introduction}

Time-series data is frequently used in various real-world systems, especially in multivariate scenarios such as server machines, water treatment plants, spacecraft, etc. Detecting an anomalous event in such time-series data is crucial to managing those systems~\cite{su2019smdomnianomaly,mathur2016swatwadi,hundman2018smapmsl,6684530,blazquez2021review}. To solve this problem, several classical approaches have been developed in the past~\cite{fox1972outliers,zhang2005network,ma2003time,liu2008isolationforest}. However, due to the limited capacity of their approaches, they could not fully capture complex, non-linear, and high-dimensional patterns in the time-series data. 

Recently, various unsupervised approaches employing deep learning architectures have been proposed. Such works include adopting architectures such as recurrent neural networks (RNN)~\cite{hundman2018smapmsl}, variational autoencoders (VAE)~\cite{xu2018unsupervised}, generative adversarial networks (GAN)~\cite{madgan}, graph neural networks (GNN)~\cite{gdn}, and combined architectures~\cite{zong2018dagmm,su2019smdomnianomaly,shen2020thoc,audibert2020usad,park2018lstmvae}. These deep learning approaches have brought significant performance improvements in time-series anomaly detection. However, most deep learning-based methods have shown several downsides. First, they require a long training time due to complex calculations, hindering applications where fast and efficient training is needed. Second, they need a significant amount of effort to tune model hyperparameters (e.g., sliding window size) for a given dataset, which can be costly in real-world applications.

\begin{figure}[t]
\centering
\includegraphics[width=0.98\columnwidth]{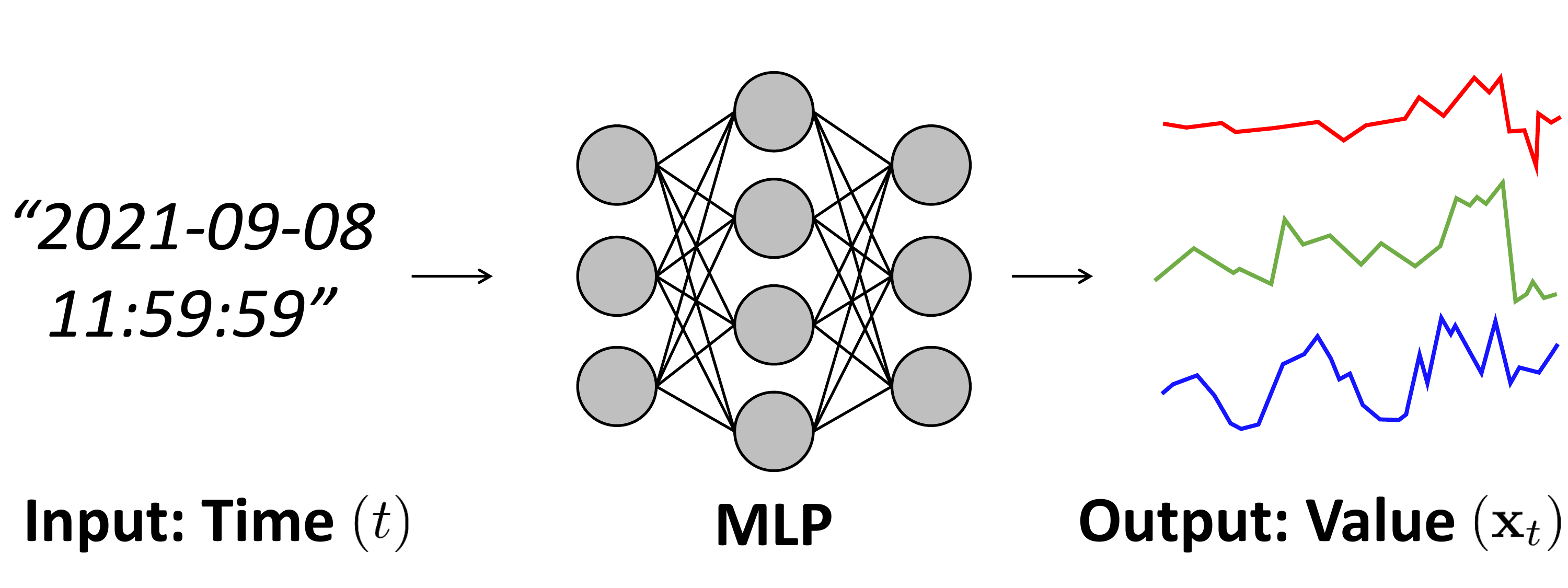} 
\caption{Implicit neural representation for multivariate time-series data.}
\label{Figure1INRinTimeseries}
\end{figure}

In our paper, we propose Implicit Neural Representation-based Anomaly Detection (INRAD), a novel approach that performs anomaly detection in multivariate time-series data by adopting implicit neural representation (INR). Figure~\ref{Figure1INRinTimeseries} illustrates the approach of INR in the context of multivariate time-series data. Unlike conventional approaches where the values are passed as input to the model (usually processed via sliding window etc.), we directly input time to a multi-layer perceptron (MLP) model. Then the model tries to represent the values of that time, which is done by minimizing a mean-squared loss between the model output and the ground truth values. In other words, we train a MLP model to represent the time-series data itself. Based on our observation that the INR represents abnormal data relatively poorly compared to normal data, we use the representation error as the anomaly score for anomaly detection. Adopting such a simple architecture design using MLP naturally results in a fast training time. Additionally, we propose a temporal encoding technique that improves efficiency for the model to represent time-series data, resulting in faster convergence time. 

In summary, the main contributions of our work are:
 
\begin{itemize}
    \item We propose INRAD, a novel time-series anomaly detection method that only uses a simple MLP which maps time into its corresponding value.
    \item We introduce a temporal encoding technique to represent time-series data efficiently.
    \item We conduct extensive experiments while using the same set of hyperparameters over all five real-world benchmark datasets. Our experimental results show that our proposed method outperforms previous state-of-the-art methods in terms of not only accuracy, but also training speed in a highly robust manner.
\end{itemize}

\section{Related Work}

In this section, we review previous works for time-series anomaly detection and implicit neural representation. 

\subsection{Time-Series Anomaly Detection} 

Since the first study on this topic was conducted by \cite{fox1972outliers}, time-series anomaly detection has been a topic of interest over the past decades~\cite{6684530,blazquez2021review}. Traditionally, various methods, including autoregressive moving average (ARMA)~\cite{galeano2006outlier} and autoregressive integrated moving average (ARIMA) model~\cite{zhang2005network}-based approaches, one-class support vector machine-based method~\cite{ma2003time} and isolation-based method \cite{liu2008isolationforest} have been widely introduced for time-series anomaly detection. However, these classical methods either fail to capture complex and non-linear temporal characteristics or are very sensitive to noise, making them infeasible to be applied on real-world datasets.

Recently, various unsupervised deep learning-based approaches have successfully improved performance in complex multivariate time-series anomaly detection tasks. As one of the well-known unsupervised models, autoencoder (AE)-based approaches~\cite{sakurada2014anomaly} capture the non-linearity between variables. Recurrent neural networks (RNNs) are a popular architecture choice used in various methods~\cite{hundman2018smapmsl,malhotra2016lstm} for capturing temporal dynamics of time series data. Generative models are also used in the literature, namely generative adversarial networks~\cite{madgan} and variational autoencoder (VAE)-based approaches~\cite{xu2018unsupervised}. Graph neural network-based approach~\cite{gdn} is also proposed to capture the complex relationship between variables in the multivariate setting. Furthermore, methodologies combining multiple architectures are also proposed, such as AE with the Gaussian mixture model~\cite{zong2018dagmm} or AE with GANs~\cite{audibert2020usad}, stochastic RNN with a planar normalizing flow~\cite{su2019smdomnianomaly}, deep support vector data description~\cite{deepsvdd} with dilated RNN~\cite{drnn}, and VAE with long short term memory (LSTM) networks~\cite{park2018lstmvae}.
    
Despite remarkable improvements via those above deep learning-based approaches, most of the approaches produce good results at the expense of training speed and generalizability. Such long training time with costly hyperparameter tuning for each dataset results in difficulties applying to practical scenarios~\cite{audibert2020usad}. 

\subsection{Implicit Neural Representation} 

Recently, implicit neural representations (or coordinate-based representations) have gained popularity, mainly in 3D deep learning. Generally, it trains a MLP to represent a single data instance by mapping the coordinate (e.g., $xyz$-coordinates) to the corresponding values of the data. This approach has been proven to have expressive representation capability with memory efficiency. As one of the well-known approaches, occupancy networks~\cite{mescheder2019occunet} train a binary classifier to predict whether a point is inside or outside the data to represent. DeepSDF~\cite{park2019deepsdf} directly regresses a signed distance function that returns a signed distance to the closest surface when the position of a 3D point is given. Instead of occupancy networks or signed distance functions, NeRF~\cite{mildenhall2020nerf} proposes to map an MLP to the color and density of the scene to represent. SIREN~\cite{sitzmann2020siren} proposes using sinusoidal activation functions in MLPs to facilitate high-resolution representations. Since then, various applications, including view synthesis~\cite{martin2021nerfw} and object appearance~\cite{saito2019pifu} have been widely studied.

However, the application of INR to time-series data has been relatively underdeveloped. Representation of time-varying 3D geometry has been explored~\cite{niemeyer2019occupancyflow}, but they do not investigate multivariate time-series data. Although SIREN~\cite{sitzmann2020siren} showed the capability to represent audio, its focus was limited to the high-quality representation of the input signals. To the best of our knowledge, this is the first work to use INR to solve the problem of time-series anomaly detection.

\section{INRAD Framework}

In this section, we define the problem that we aim to solve, and then we present our proposed INRAD based on the architecture proposed by \cite{sitzmann2020siren}. Next, we describe our newly designed temporal encoding technique in detail. Finally, we describe the loss function to make our model represent input time signals and describe the anomaly score used during the detection procedure. Figure~\ref{Figure2Overview} describes the overview of the proposed method.

\begin{figure*}[t]
\centering
\includegraphics[width=0.9\textwidth]{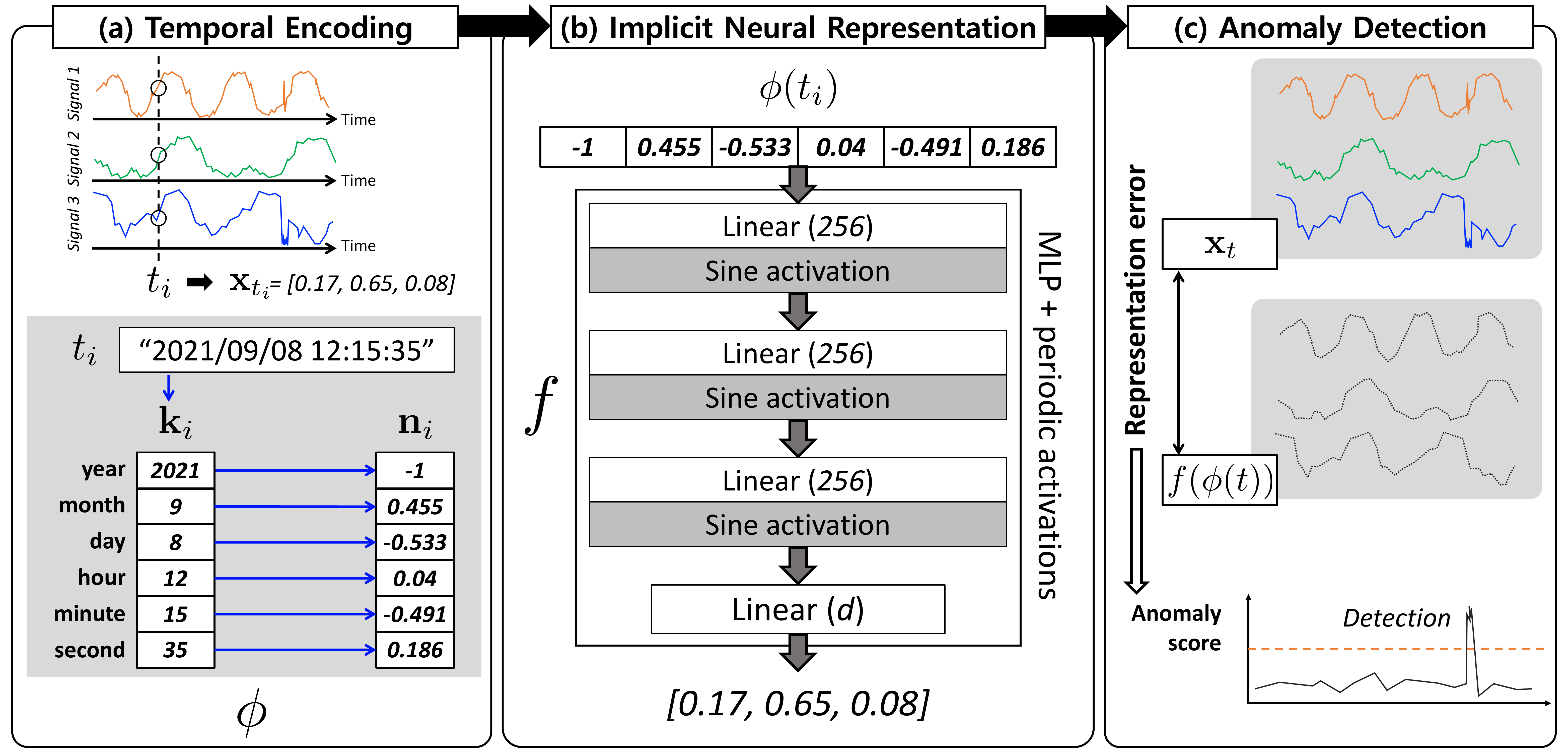}
\caption{The overview of the proposed Implicit Neural Representation-based Anomaly Detection (INRAD). (a) From the given time-series data, we perform temporal encoding and represent time as a real-valued vector. (b) An MLP using periodic activation functions represents the given data by mapping the time processed by temporal encoding to the corresponding values. (c) After the model converges, we calculate the representation error and use this as the anomaly score for detection.}
\end{figure*}\label{Figure2Overview}

\subsection{Problem Statement}
In this section, we formally state the problem of multivariate time-series anomaly detection as follows.

We first denote multivariate time-series data as $X = \{({t_1}, \mathbf{x}_{t_1}), ({t_2}, \mathbf{x}_{t_2}), ({t_3}, \mathbf{x}_{t_3}),...,({t_N}, \mathbf{x}_{t_N})\}$, where $t_i$ denotes a timestamp, $\mathbf{x}_{t_i}$ denotes corresponding values at the timestamp, and $N$ denotes the number of observed values. As we focus on multivariate data, $\mathbf{x}_{t_{i}}$ is a $d$-dimensional vector representing multiple signals. The goal of time-series anomaly detection is to output a sequence, $Y = \{y_{t_1}, y_{t_2}, y_{t_3},...,y_{t_N}\}$, where $y_{t_i} \in \{0, 1\}$ denotes the abnormal or normal status at $t_i$. In general, 1 indicates the abnormal state while 0 indicates normal state.


\subsection{Implicit Neural Representation of Time-Series Data}
To represent a given time-series data, we adopt the architecture proposed by \cite{sitzmann2020siren}, which leverages periodic functions as activation functions in the MLP model, resulting in a simple yet powerful model capable of representing various signals, including audio. After preprocessing the time coordinate input via an encoding function $\phi$, our aim is to learn a function $f$ that maps the encoded time $\phi(t_i)$ to its corresponding value $\mathbf{x}_{t_i}$ of the data.

We can describe the MLP $f$ by first describing each fully-connected layer and stacking those layers to get the final architecture. Formally, the $l$th fully-connected layer $f_{l}$ with hidden dimension $m_l$ can be generally described as $f_{l}(\mathbf{h}_{l-1}) = \sigma(\mathbf{W}_{l}\mathbf{h}_{l-1} + \mathbf{b}_l)$, where $\mathbf{h}_{l-1} \in \mathbb{R}^{m_{l-1}}$ represents the output of the previous layer $f_{l-1}$, $\mathbf{W}_{l} \in \mathbb{R}^{m_{l} \times m_{l-1}}$ and $\mathbf{b}_{l} \in \mathbb{R}^{m_{l}}$ are learnable weights and biases, respectively, and $\sigma$ is a non-linear activation function. Here, sine functions are used as $\sigma$, which enables accurate representation capabilities of various signals. In practice, a scalar $\omega_0$ is multiplied such that the $l$th layer is $f_l = \sin (\omega_0 \cdot \mathbf{W}_{l}\mathbf{h}_{l-1} + \mathbf{b}_l)$, in order for the input to span multiple periods of the sine function. 

Finally, by stacking a total of $L$ layers with an additional linear transformation at the end, we now have our model $f(\phi(t_i)) = \mathbf{W}(f_{L} \circ f_{L-1} \circ \cdots \circ f_1)(\phi(t_i)) + \mathbf{b}$ which maps the input $t_i$ to the output $f(\phi(t_i)) \in \mathbb{R}^{d}$.

\subsection{Temporal Encoding}\label{temporal encoding}

As INR has been mainly developed to represent 2D or 3D graphical data, encoding time coordinate for INR has rarely been studied. Compared to graphical data, which the number of points in each dimension is fairly limited to around thousands, the number of timestamps is generally much larger and varies among different datasets. Also, training and test data need to be considered regarding their chronological order (training data usually comes first). These observations with a real-world time-series data motivate us to design a new encoding strategy such that 1) the difference between $\phi(t_i)$ and $\phi(t_{i+1})$ is not affected by the length of the time sequence 2) chronological order between train and test data is preserved after encoding 3) timestamps from real-world data are naturally represented using its standard time scale rather than relying on the sequential index of time-series data. These desired properties are not satisfied with the encoding strategy applied in~\cite{sitzmann2020siren} (which we call vanilla encoding), where it normalizes coordinates in the range $[-1, 1]$.

We now describe our temporal encoding, a simple yet effective method which satisfies conditions mentioned above. The key idea is to directly utilize the timestamp data present in the time-series data (we can assign arbitrary timestamps if none is given). \color{black} We first represent $t_i$ into a 6-dimensional vector ${\bf k} = [k_{yr}, k_{mon}, k_{day}, k_{hr}, k_{min}, k_{sec}] \in \mathbb{R}^{6}$, each dimension representing year, month, day, hour, minute, and second respectively. Here, $k_{yr},k_{mon},k_{day},k_{hr},k_{min}$ are all positive integers, while $k_{sec} \in [0,60)$. Note that this can flexibly change depending on the dataset. For instance, if the timestamp does not include minute and second information, we use a 4-dimensional vector (i.e., $[k_{yr},k_{mon},k_{day},k_{hr}]$ and $k_{hr} \in [0, 24)$).

Next, we normalize the vectorized time information. With a pre-defined year $k'_{yr}$, we first set $[k'_{yr},1,1,0,0,0]$ (January 1st 00:00:00 at year $k'_{yr}$) as [-1,-1,-1,-1,-1,-1]. Now, let us represent the current timestamp of interest as $\mathbf{k}^{curr}$. We normalize the $j$-th dimension of the current timestamp $\mathbf{k}^{curr}$ by the following linear equation:
\begin{equation}\label{temporalencoding}
    n_j^{curr} = -1 + \dfrac{1 - (-1)}{N_{j}-1} \times (k_j^{curr}-\mathbb{I}(j=1)k'_{yr})
\end{equation}
where $n_i^{curr}$ is the $j$th dimension of the normalized vector $\mathbf{n}^{curr} \in \mathbb{R}^{6}$ and $\mathbb{I}$ is an indicator function. For the values of $N_i$, we set $N_2 = 12, N_3 = 31, N_4 = 24, N_5 = 60, N_6 = 60$ to match the standard clock system. We assume that $N_1$ is pre-defined by the user. In short, we define a temporal encoding function $\phi$ that transforms a scalar $t$ into $\mathbf{n}_i$ ($\phi: t_i \mapsto \mathbf{n}_i$). In our method, we will by default use this temporal encoding unless otherwise stated. 

\subsection{Loss Function}
As we aim the model to represent the input time-series data, we compare the predicted value at each timestamp $t_i$to its ground-truth value $\mathbf{x}_{t_i}$. Therefore, we minimize the following loss function: 
\begin{equation}
    \mathcal{L} = \dfrac{1}{n} \sum_{i=1}^{n} ||\mathbf{x}_{t_{i}} - f({\phi}(t_{i}))||^{2}
\end{equation}
where $||\cdot||$ indicates the l2 norm of a vector.

\subsection{Anomaly Score and Detection Procedure}

Our proposed representation error-based anomaly detection strategy is built on the observation that values at an anomalous time are difficult to represent, resulting in relatively high representation error. By our approach described above, the given data sequence $X$ is represented by an MLP function $f$. We now perform anomaly detection with this functional representation by defining the representation error as the anomaly score. Formally, the anomaly score $a_{t_i}$ at a specific timestamp $t_i$ is defined as $a_{t_i} = |\mathbf{x}_{t_i} - f({\phi}(t_i))|$, where $|\cdot|$ indicates the l1 norm of a vector. Anomalies can be detected by comparing the anomaly score $a_{t_i}$ with the pre-defined threshold $\tau$.

In our approach, we first use the training data to pre-train our model $f$ and then re-train the model to represent the given test data to obtain the representation error as an anomaly score for the detection.

\section{Experiments}
In this section, we perform various experiments to answer the following research questions:
\begin{itemize}
    \item {\bf RQ1:} Does our method outperform various state-of-the-art methods, even with a fixed hyperparameter setting?
    \item {\bf RQ2:} How does our proposed temporal encoding affect the performance and convergence time?
    \item {\bf RQ3:} Does our method outperform various state-of-the-art methods in terms of training speed?
    \item {\bf RQ4:} How does our method behave in different hyperparameter settings?
\end{itemize}


\begin{table}[t]
\centering
\begin{tabular}{r|c|c|c|c} \toprule
     Datasets & Train & Test & Features & Anomalies \\\midrule
    SMD &  708405 & 708420  &  28$\times$38 &  4.16 (\%)\\
    SMAP &  135183 &  427617 &  55$\times$25  &  13.13 (\%)\\
    MSL &  58317 &  73729 &  27$\times$55 & 10.72 (\%)\\
    SWaT &  496800 &  449919 &  51 & 11.98 (\%)\\
    WADI & 1048571  &  172801 &  123  & 5.99 (\%)\\
    \bottomrule
\end{tabular}
\caption{Statistics of the datasets used in our experiments.}
\label{TableDatasetStatistics}
\end{table}


\subsection{Dataset}
We use five real-world benchmark datasets, SMD~\cite{su2019smdomnianomaly}, SMAP \& MSL~\cite{hundman2018smapmsl}, SWaT \& WADI~\cite{mathur2016swatwadi}, for anomaly detection for multivariate time-series data, which contain ground-truth anomalies as labels. Table \ref{TableDatasetStatistics} summarizes the statistics of each dataset, which we further describe its detail in the supplementary material.

In our experiments, we directly use timestamps included in the dataset for SWAT and WADI. We arbitrarily assign timestamps for the other three datasets since no timestamps representing actual-time information are given.

\begin{table*}[t]
\centering
\fontsize{9}{10}\selectfont
\begin{tabular}{r|ccc|ccc|ccc|ccc|ccc} \toprule
     & \multicolumn{3}{c}{SMD} & \multicolumn{3}{c}{SMAP} & \multicolumn{3}{c}{MSL} & \multicolumn{3}{c}{SWaT} & \multicolumn{3}{c}{WADI}\\\cmidrule{2-16}
    Method & P & R & F1 & P & R & F1 & P & R & F1 & P & R & F1 & P & R & F1\\ \midrule
    IF  & 59.4 & 85.3 & 0.70 & 44.2 & 51.1 & 0.47 & 56.8 & 67.4 & 0.62 & 96.2 & 73.1 & 0.83 & 62.4 & 61.5 & 0.62\\
    LSTM-VAE  & 87.0 & 78.8 & 0.83 & 71.6 & 98.8 & 0.83 & 86.0 & 97.6 & 0.91 & 71.2 & 92.6 & 0.80 & 46.3 & 32.2 & 0.38 \\
    DAGMM  & 67.3 & 84.5 & 0.75 & 63.3 & 99.8 & 0.78 & 75.6 & 98.0 & 0.85 & 82.9 & 76.7 & 0.80 & 22.3 & 19.8 & 0.21\\
    OmniAnomaly  & 98.1 & 94.4 & 0.96 & 75.9 & 97.6 & 0.85 & 91.4 & 88.9 & 0.90 & 72.2 & 98.3 & 0.83 & 26.5 & 98.0 & 0.41\\
    USAD  & 93.1 & 94.4 & 0.96 & 77.0 & 98.3 & 0.86 & 88.1 & 97.9 & 0.93 & 98.7 & 74.0 & 0.85 & 64.5 & 32.2 & 0.43\\
    THOC  & 73.2 & 78.8 & 0.76 & 79.2 & 99.0 & 0.88 & 78.9 & 97.4 & 0.87 & 98.0 & 70.6 & 0.82 & - & - & - \\ \midrule
    $\text{INRAD}^{\text{c}}_{\text{van}}$ & 94.7 & 97.8 & {0.96} & 80.0 & 99.3 & 0.89 & 93.6 & 98.1 & {\bf 0.96} & 96.9 & 88.7 & 0.93 & 60.2 & 67.0 & 0.63 \\
    $\text{INRAD}^{\text{c}}_{\text{temp}}$ & 98.0 & 98.3 & {\bf 0.98} & 83.2 & 99.1 & 0.90 & 92.1 & 99.0 & 0.95 & 93.0 & 96.3 & 0.95 & 78.4 & 99.9 & 0.88 \\ \midrule
    $\text{INRAD}_{\text{van}}$  & 98.0 & 98.6 & {\bf 0.98} & 84.0 & 99.4 & {0.91} & 90.4 & 99.0 & {0.95} & 96.4 & 91.7 & {0.94} & 77.1 & 66.5 & {0.71} \\ 
    $\text{INRAD}_{\text{van}^{*}}$  & 95.0 & 96.4 & {0.95} & 82.6 & 99.3 & {0.90} & 91.7 & 98.7 & {0.95} & 84.2 & 84.7 & {0.84} & 72.4 & 72.8 & {0.73} \\ 
    $\text{INRAD}_{\text{temp}}$  & 98.2 & 97.5 & {\bf 0.98} & 85.8 & 99.5 & {\bf 0.92} & 93.3 & 99.0 & {\bf 0.96} & 95.6 & 98.8 & {\bf 0.97} & 88.9 & 99.1 & {\bf 0.94} \\ \bottomrule
\end{tabular}
\caption{Anomaly detection accuracy results in terms of precision(\%), recall(\%), and F1-score, on five real-world benchmark datasets. $\text{INRAD}_{\text{van}}$, $\text{INRAD}_{\text{van}^{*}}$, and $\text{INRAD}_{\text{temp}}$ adopts the vanilla, vanilla$^{*}$, and temporal encoding, respectively. Also, $\text{INRAD}^{\text{c}}_{\text{van}}$ and $\text{INRAD}^{\text{c}}_{\text{temp}}$ indicates that the experiment was run on the cold-start setting with each encoding.}
\label{TablePerformanceComparison}
\end{table*}

\subsection{Baseline methods}
We demonstrate the performance of our proposed method, INRAD, by comparing with the following six anomaly detection methods:

\begin{itemize}
    \item {\bf IF}~\cite{liu2008isolationforest}: Isolation forests (IF) is the most well-known isolation-based anomaly detection method, which focuses on isolating abnormal instances rather than profiling normal instances. 
    \item {\bf LSTM-VAE}~\cite{park2018lstmvae}: LSTM-VAE uses a series of connected variational autoencoders and long-short-term-memory layers for anomaly detection.
    \item {\bf DAGMM}~\cite{zong2018dagmm}: DAGMM is an unsupervised anomaly detection model which utilizes an autoencoder and the Gaussian mixture model in an end-to-end training manner. 
    \item {\bf OmniAnomaly}~\cite{su2019smdomnianomaly}: OmniAnomaly employs a stochastic recurrent neural network for multivariate time-series anomaly detection to learn robust representations with a stochastic variable connection and planar normalizing flow.
    \item {\bf USAD}~\cite{audibert2020usad}: USAD utilizes an encoder-decoder architecture with an adversely training framework inspired by generative adversarial networks.
    \item {\bf THOC}~\cite{shen2020thoc}: THOC combines a dilated recurrent neural network~\cite{drnn} for extracting multi-scale temporal features with the deep support vector data description~\cite{deepsvdd}. 
\end{itemize}

\subsection{Evaluation Metrics}

We use precision (P), recall (R), F1-score (F1) for evaluating time-series anomaly detection methods. Since these performance measures depend on the way threshold is set on the anomaly scores, previous works proposed a strategy such as applying extreme value theory~\cite{siffer2017anomaly}, using a dynamic error over a time window \cite{hundman2018smapmsl}. However, not all methodologies develop a mechanism to select a threshold in different settings, and many previous works~\cite{audibert2020usad,su2019smdomnianomaly,xu2018unsupervised} adopt the best F1 score for performance comparison, where the optimal global threshold is chosen by trying out all possible thresholds on detection results. We also use the point-adjust approach~\cite{xu2018unsupervised}, widely used in evaluation~\cite{audibert2020usad,su2019smdomnianomaly,shen2020thoc}. Specifically, if any point in an anomalous segment is correctly detected, other observations in the segment inside the ground truth are also regarded as correctly detected.

Therefore, we adopt the best F1-score (short F1 score hereafter) and the point-adjust approach for evaluating the anomaly detection performance to directly compare with the aforementioned state-of-the-art methods. 

\subsection{Hyperparameters and Experimental Setup}

To show the robustness of our proposed method, we conduct experiments using the same hyperparameter setting for all benchmark datasets. 

The details of the experimental setting are as follows. For the model architecture, we use a 3-layer MLP with sinusoidal activations with 256 hidden dimensions each (refer to Figure~\ref{Figure2Overview}(b)). Following~\cite{sitzmann2020siren}, we set $\omega_0 = 30$ except for the first layer, which is set to $3000$. During training, we use the Adam optimizer~\cite{kingma2014adam} with a learning rate of 0.0001 and $(\beta_1, \beta_2) = (0.9, 0.99)$. Additionally, we use early stopping with patience 30. Our code and data are released at https://github.com/KyeongJoong/INRAD

\subsection{RQ 1. Performance Comparison}

Table~\ref{TablePerformanceComparison} shows the performance comparison results of our proposed method $\text{INRAD}_{\text{temp}}$ and its variants, along with other baseline approaches on five benchmark datasets. We use the reported accuracy values of baselines (except THOC~\cite{shen2020thoc}) from the previous work \cite{audibert2020usad}, which achieves state-of-the-art performance in the identical experimental setting with ours, such as datasets, train/test split, and evaluation metrics. Note that results of THOC on the WADI dataset are omitted due to an out-of-memory issue. 

Overall, our proposed $\text{INRAD}_{\text{temp}}$ consistently achieves the highest F1 scores over all datasets. Especially, the performance improvement over the next best method achieves 0.32 in terms of F1 score on the WADI dataset, where most other approaches show relatively low performance. On other datasets, we still outperform the second-best performance of other baselines by 0.02 to 0.12. Considering that the single hyperparameter setting restriction is only applied for our method, this shows that our approach can provide superior performance in a highly robust manner in various datasets.


As we adopt the representation error-based detection strategy, it is possible that our method detects anomalies without training data by directly representing the test set. We hypothesize that the test set already contains an overwhelming portion of normal samples from which the model can still learn temporal dynamics of normal patterns in the given data without any complex model architectures (e.g., RNN and its variants). To distinguish from the original method, we denote this variant as $\text{INRAD}^{\text{c}}_{\phi}$ ($\phi=$ 'van' or 'temp' for vanilla and temporal encoding, respectively), which we also investigate its performance. We observe that both $\text{INRAD}^{\text{c}}_{\text{van}}$ and $\text{INRAD}^{\text{c}}_{\text{temp}}$ generally achieves slightly inferior performance to $\text{INRAD}_{\text{temp}}$ except for WADI. This result shows that utilizing the training dataset has performance benefits, especially when the training data is much longer than the test data. 



\begin{table}[t]
\centering
\fontsize{9}{10}\selectfont
\begin{tabular}{r|c|c|c} \toprule
    Method & SMD & SMAP & MSL  \\ \midrule
    LSTM-VAE  & 3.807 & 0.987 & 0.674  \\
    OmniAnomaly  & 77.32 & 16.66 & 15.55  \\
    USAD  & 0.278 & 0.034 & 0.029   \\
    THOC  & 0.299 & 0.07 & 0.066  \\\midrule
    INRAD  & 0.243 & 0.024 & 0.020  \\ \bottomrule
\end{tabular}
\caption{Comparison of training time (sec) per epoch.}
\label{TableTrainingTimePerEpoch}
\end{table}

\subsection{RQ 2. Effect of temporal encoding}

We study the effect of our temporal encoding method by comparing it with two encoding methods: vanilla and its variant, vanilla$^{*}$. Vanilla encoding normalizes the indices $[1,2, \cdots, M]$ of the training data to $[-1, 1]$, and the indices of the test data are also mapped to $[-1, 1]$. On the other hand, vanilla$^{*}$ encoding is derived from vanilla encoding to preserve chronological order and unit interval of training and test data after encoding by mapping indices of test data to the range $[1,\infty)$, while keeping the difference between neighboring encoding consistent with the training data. We denote the variant using vanilla and vanilla$^{*}$ as $\text{INRAD}_{\text{van}}$ and $\text{INRAD}_{\text{van}^{*}}$, respectively.

Table~\ref{TablePerformanceComparison} shows that $\text{INRAD}_{\text{temp}}$ achieves slightly superior performance in general compared to $\text{INRAD}_{\text{van}}$ and $\text{INRAD}_{\text{van}^{*}}$. However, when the length of the dataset becomes long, the performance of vanilla and vanilla$^{*}$ degrades significantly while temporal encoding remains at 0.94, as shown in the case of WADI. The performance gap in WADI becomes even more significant in the case of cold-start settings, which is 0.25.

\begin{figure}[t]
\centering
\includegraphics[width=0.9\columnwidth]{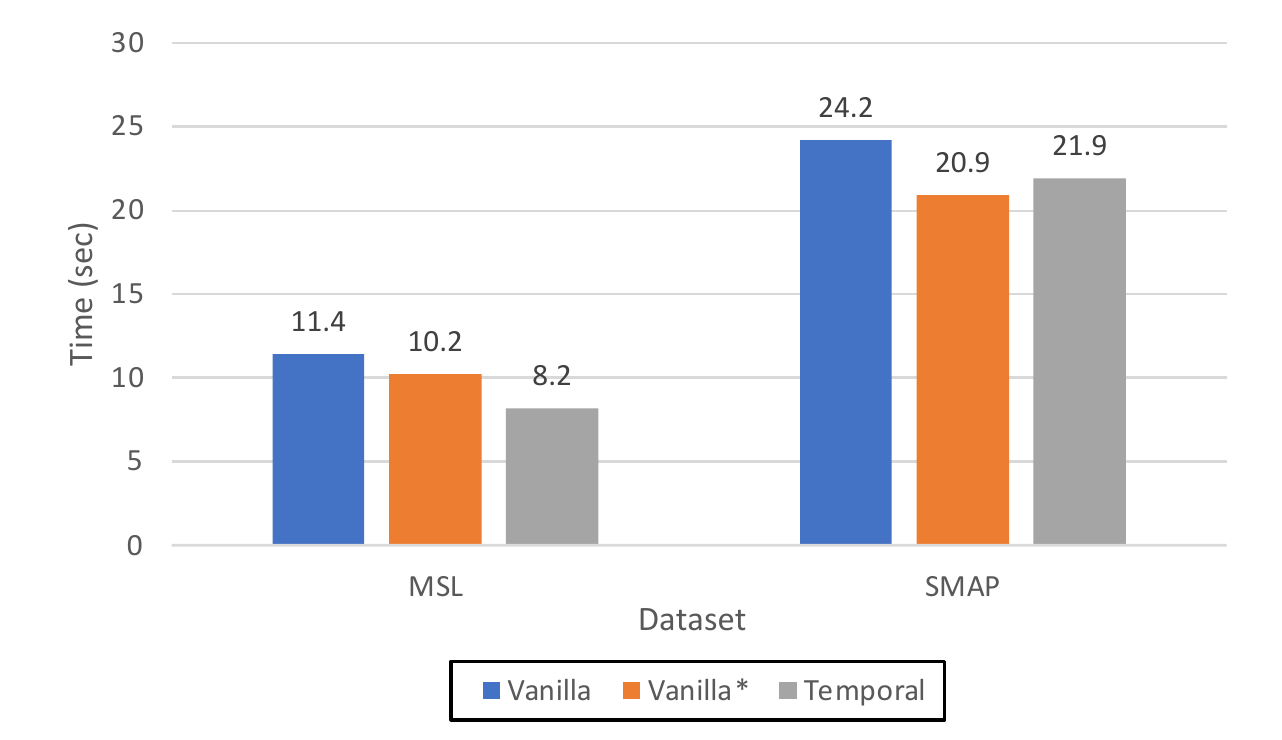} 
\caption{Convergence time (sec) comparison for different encoding techniques.}
\label{Figure3Encoding}
\end{figure}

Also, Figure~\ref{Figure3Encoding} compares the convergence time for representation of test data between $\text{INRAD}_{\text{van}}$, $\text{INRAD}_{\text{van}^{*}}$, and $\text{INRAD}_{\text{temp}}$ using MSL and SMAP dataset. The vanilla encoding shows the slowest convergence time, and the vanilla$^{*}$ and our temporal encoding shows competitive results. This result shows that representation of test data is learned faster when time coordinates in training and test data are encoded while preserving the chronological order. Overall, our temporal encoding strategy achieves superior performance and fast convergence compared to the vanilla encoding strategy. 

\subsection{RQ 3. Training speed comparison}

Here, we study the training speed of $\text{INRAD}_{\text{temp}}$ and compare it to the other four baselines that show good performance. Table~\ref{TableTrainingTimePerEpoch} summarizes the results the training time per epoch for $\text{INRAD}_{\text{temp}}$ along with other baseline methods on three benchmark datasets. Specifically, the reported time is the average time across all entities within each dataset (i.e., 28 entities for SMD, 55 for SMAP, and 27 for MSL). The results show that our method achieves the fastest training time, mainly because our method only uses a simple MLP for training without any additional complex modules (e.g., RNNs).

\subsection{RQ 4. Hyperparameter sensitivity}

\begin{figure}
    \centering
    \begin{subfigure}[b]{0.475\columnwidth}
        \centering
        \includegraphics[width=\columnwidth]{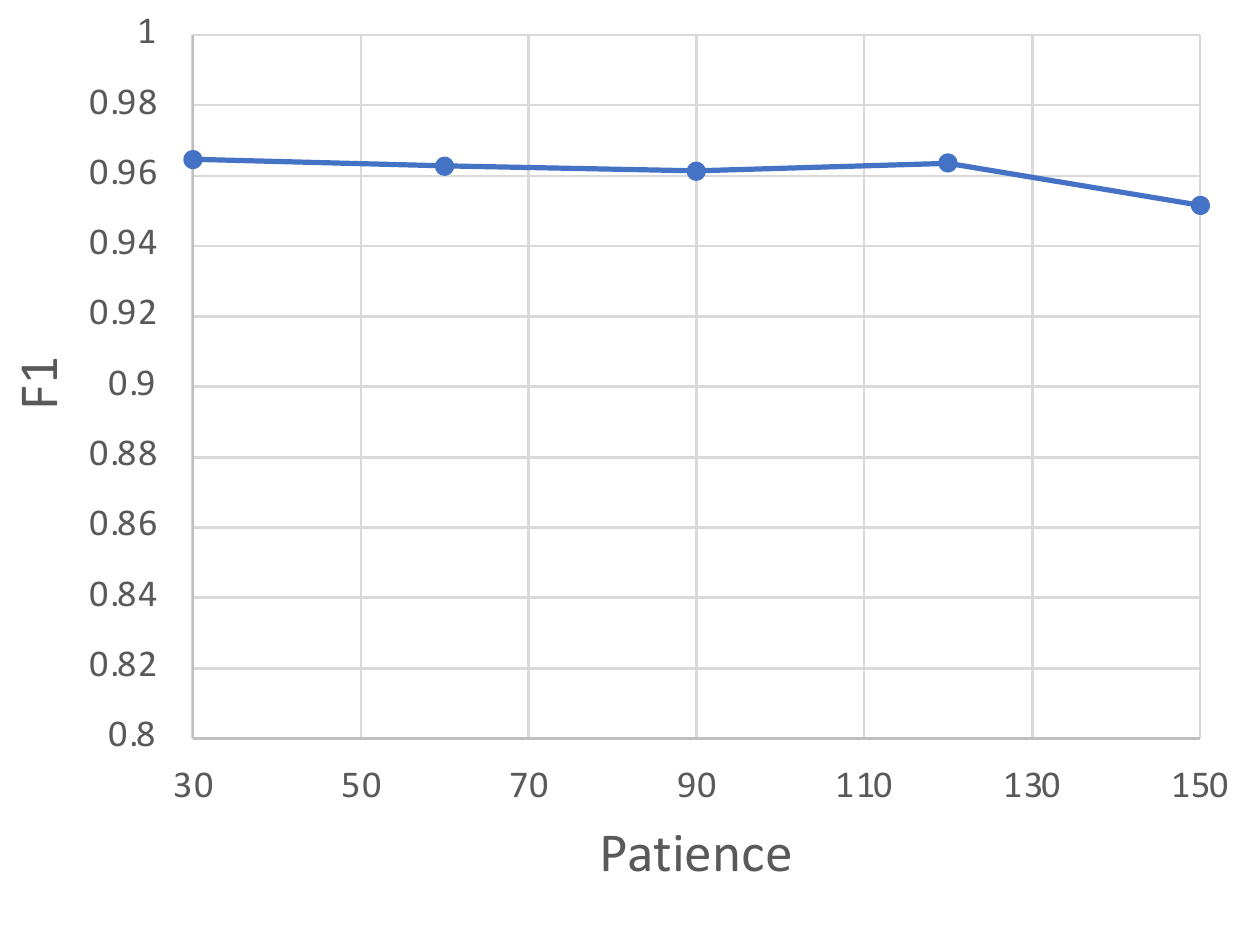}
        \caption[Network2]%
        {{\small Effect of patience}}    
        \label{Hyperparameter_patience}
    \end{subfigure}
    \hfill
    \begin{subfigure}[b]{0.475\columnwidth}
        \centering 
        \includegraphics[width=\columnwidth]{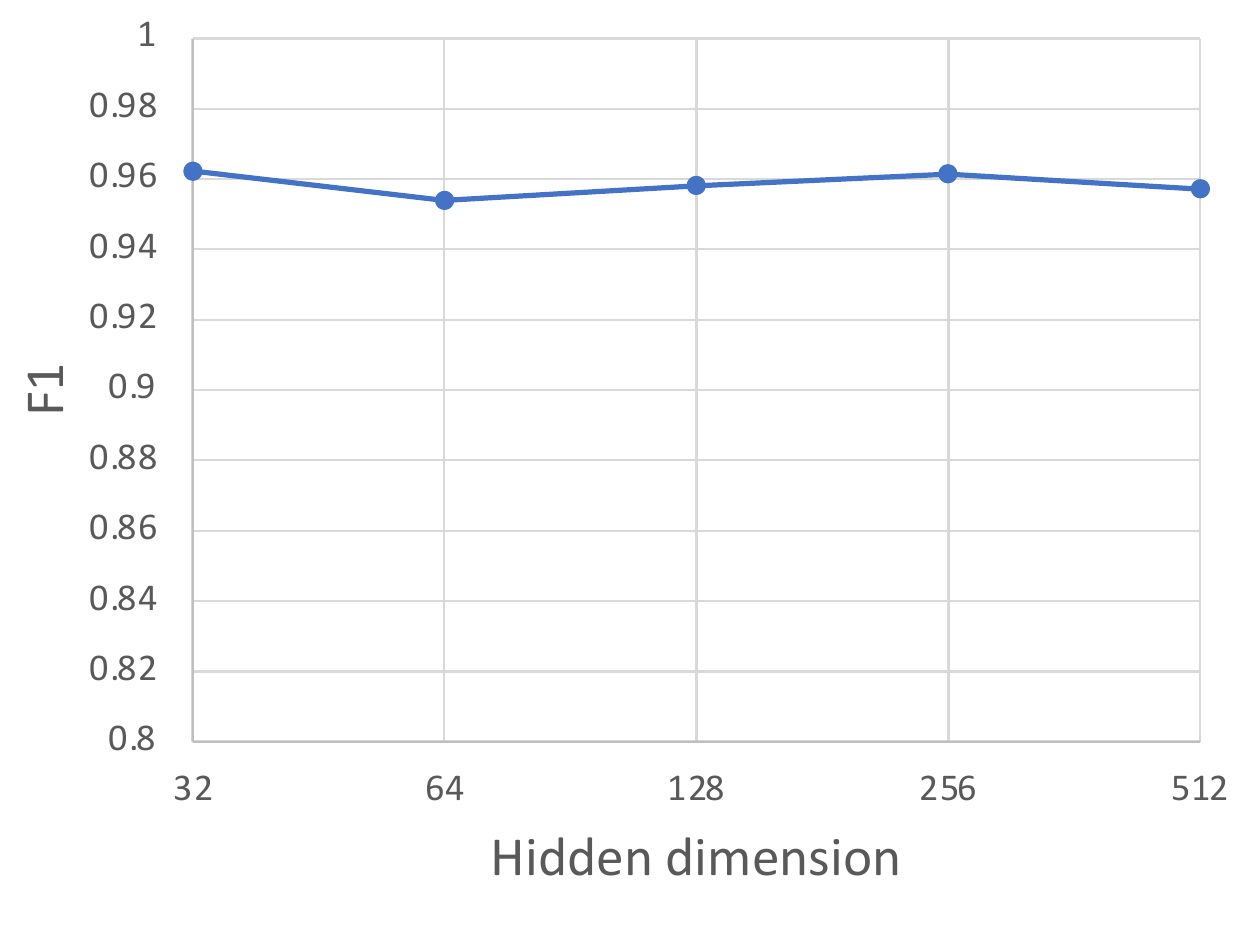}
        \caption[]%
        {{\small Effect of hidden dimension}}    
        \label{Hyperparameter_hidden_dim}
    \end{subfigure}
    \vskip\baselineskip
    \begin{subfigure}[b]{0.475\columnwidth} 
        \centering 
        \includegraphics[width=\columnwidth]{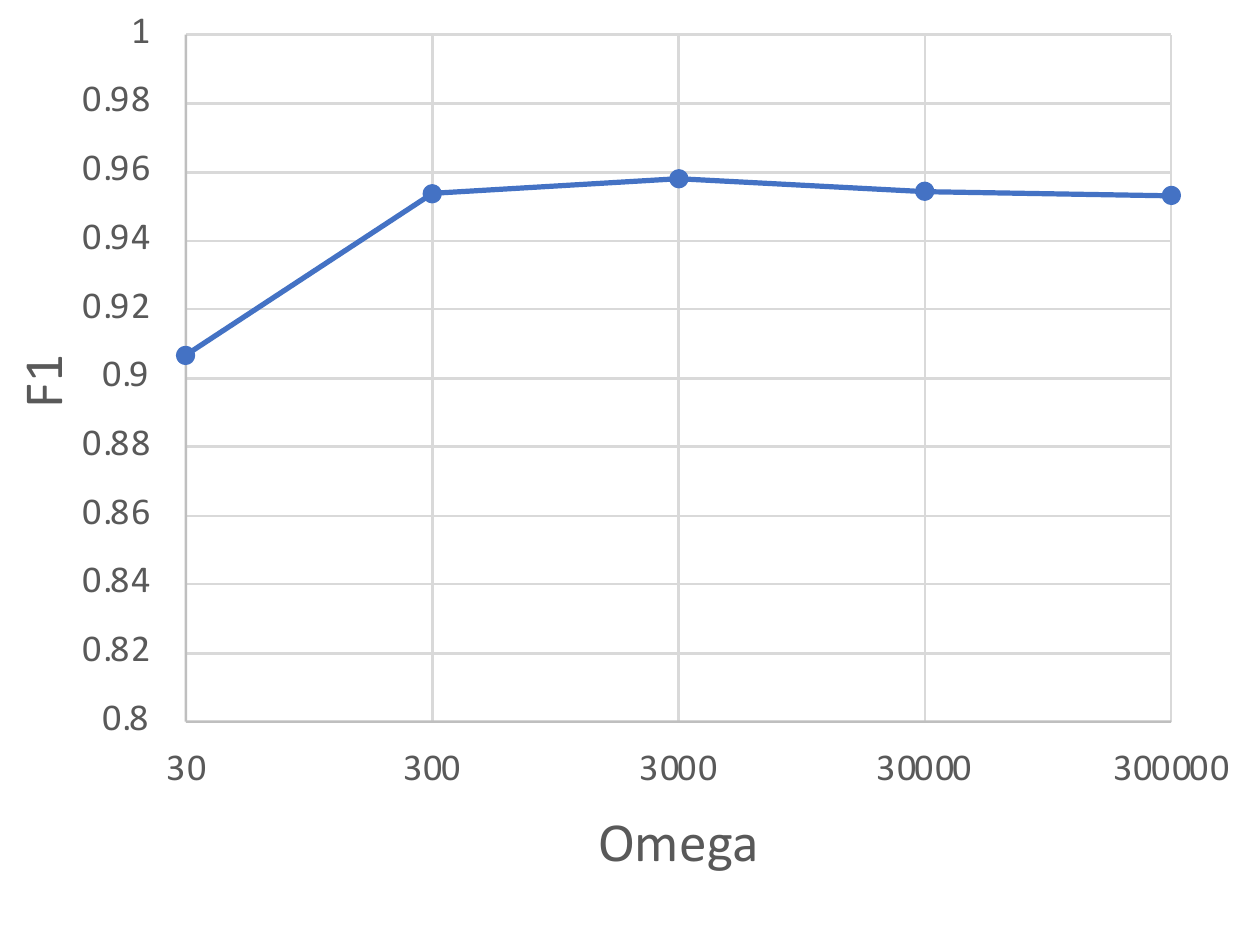}
        \caption[]%
        {{\small Effect on $\omega_0$ in first layer}}    
        \label{Hyperparameter_omega}
    \end{subfigure}
    \hfill
    \begin{subfigure}[b]{0.475\columnwidth}
        \centering 
        \includegraphics[width=\columnwidth]{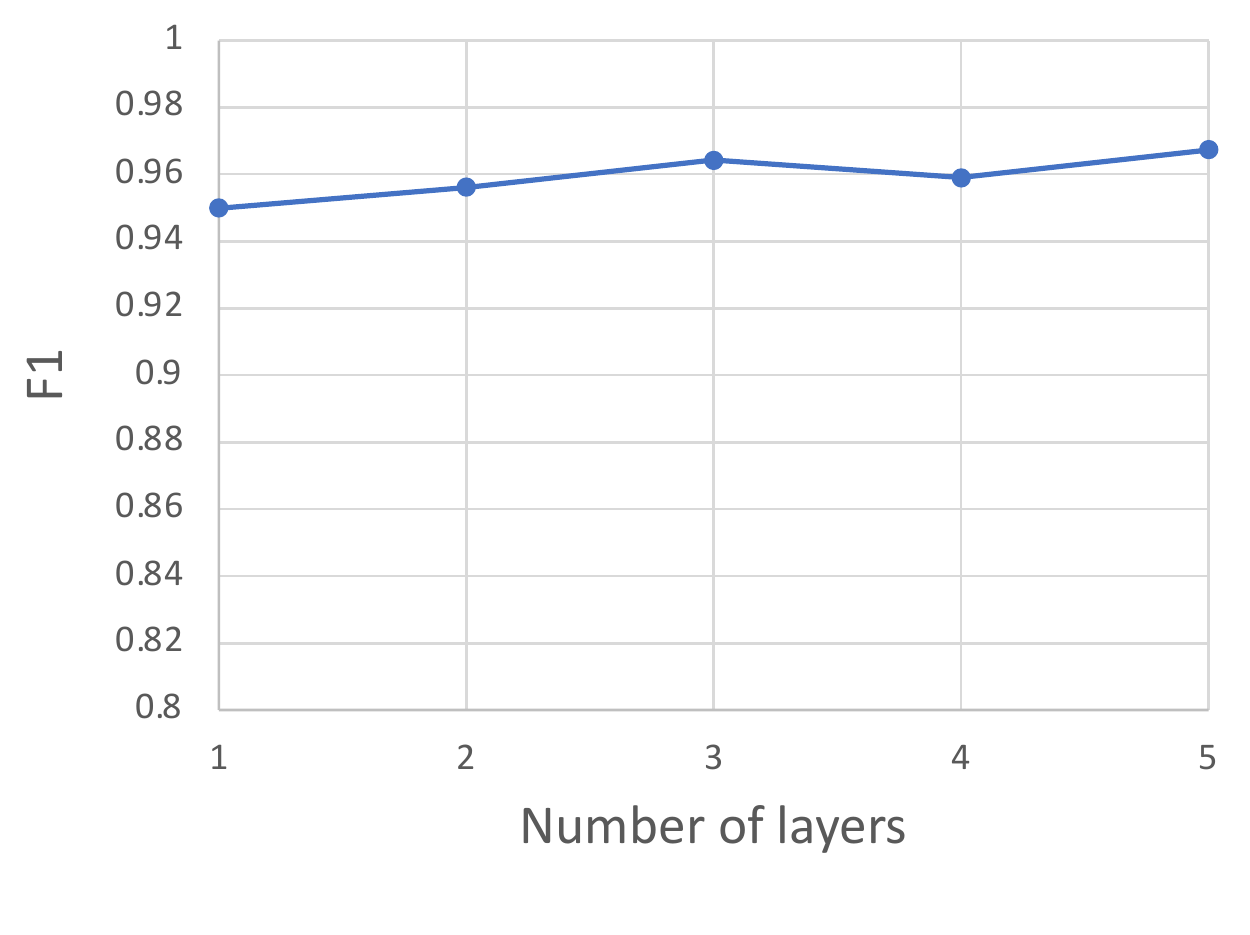}
        \caption[]%
        {{\small Effect of number of layers}}    
        \label{Hyperparameter_num_layers}
    \end{subfigure}
    \caption[ The average and standard deviation of critical parameters ]
    {\small Performance of $\text{INRAD}_{\text{temp}}$ for MSL under various hyperparameter settings.} 
    \label{Hyperparameter_experiment}
\end{figure}

In Figure~\ref{Hyperparameter_experiment}, we test the robustness of $\text{INRAD}_{\text{temp}}$ by varying different hyperparameters settings using the MSL dataset. In our experiment, we change the patience in early stopping in range $\{30, 60, 90, 120, 150\}$, size of hidden dimension in range $\{32, 64, 128, 256, 512\}$, $\omega_0$ of the first layer in range $\{30, 300, 3000, 30000, 300000\}$, and the number of layers in range $\{1, 2, 3, 4, 5\}$. Figure~\ref{Hyperparameter_patience},~\ref{Hyperparameter_hidden_dim}, and~\ref{Hyperparameter_num_layers} shows that $\text{INRAD}_{\text{temp}}$ achieves high robustness with varying hyperparameter settings. We see that the choice of $\omega_0$ also minimally impacts $\text{INRAD}_{\text{temp}}$ as in Figure~\ref{Hyperparameter_omega}. This results suggests that the MLP struggles to differentiate neighboring inputs in the case where $\omega_0$ is extremely low.

\section{Conclusion} 
In this paper, we proposed INRAD, a novel implicit neural representation-based method for multivariate time-series anomaly detection, along with a temporal encoding technique. Adopting a simple MLP, which takes time as input and outputs corresponding values to represent a given time-series data, our method detects anomalies by using the representation error as anomaly score. Various experiments on five real-world datasets show that our proposed method achieves state-of-the-art performance in terms of accuracy and training speed while using the same set of hyperparameters. For future work, we can consider additional strategies for online training in order to improve applicability in an environment where prompt anomaly detection is needed.

\nocite{langley00}

\bibliography{main}
\bibliographystyle{icml2022}

\newpage
\appendix
\onecolumn


\section{Baseline Implementation}
We describe the implementation of the baseline methods used in our paper. Isolation forest (IF)~\cite{isolationforeset} is implemented using the scikit-learn library, and we use the source code of THOC~\cite{shen2020thoc} given by authors. The other baselines are downloaded from the following links:

\begin{itemize}
    \item LSTM-VAE~\cite{park2018lstmvae}:\\ https://github.com/Danyleb/Variational-Lstm-Autoencoder
    \item DAGMM~\cite{zong2018dagmm}: \\https://github.com/tnakae/DAGMM
    \item OmniAnomaly~\cite{su2019smdomnianomaly}:\\ https://github.com/NetManAIOps/OmniAnomaly
    \item USAD~\cite{audibert2020usad}: \\https://github.com/robustml-eurecom/usad
\end{itemize}
\section{Vanilla Encoding}
In this section, we describe the encoding strategy in~\cite{sitzmann2020siren} which we call vanilla encoding in our main paper. Let us assume that the given dataset has $N$ timestamps with index $i$ (i.e., $X = \{(t_i, \mathbf{x}_{t_i})\}_{i=1}^{N}$). Also, denote the vector of indices as $\mathbf{i} = [1, 2, \cdots, N]$, and $\mathbf{1} \in \mathbb{R}^{1 \times N}$ as the one vector. Vanilla encoding plainly normalizes each indices $i$ to the range $[-1, 1]$ by $\mathbf{i}_{naive} = (2/N) \times \mathbf{i} - \mathbf{1}$.

\section{Detailed Setting of Temporal Encoding}

In the case of SWaT and WADI datasets, we use the actual timestamps given in each dataset. On the other hand, in the case of SMD, MSL, SMAP dataset that does not contain such information, we arbitrarily set the timestamps for each sample starting from "2021-01-01 00:00:00" with one-minute intervals. We assume that the test data is directly after the end of the timestamp in the training set.

Now we describe the details of the pre-defined $k'_{year}$ and $N_1$. We set $k'_{year}$ as the year of the first timestamp in the training set as we assume that our model will not encounter past information before the first sample in the training set. In the case of SWAT and WADI, we set $k'_{year}$ as 2015 and 2017, respectively, following the given timestamp information as we stated earlier. For the other datasets, we set $k'_{year}$ as 2021 as we set the first timestamp in the training set as "2021-01-01 00:00:00". Next, we set $N_1$ by $\gamma + 1$, where $\gamma$ indicates the difference between the year of the earliest and latest observed data. We note that, however, these settings can be flexibly chosen for various settings. 

\section{Detailed Description of the Datasets}
SMD~\cite{su2019smdomnianomaly} is a 5-week-long public dataset collected from a large Internet company containing data from 28 server machines, each one monitored by 33 metrics. It is divided into two subsets of equal size, where the first half is the training set and the second half is the testing set. SMAP and MSL~\cite{hundman2018smapmsl} are two real-world public datasets, expert-labeled datasets from NASA. SMAP contains the data from 55 entities monitored by 25 metrics, and MSL contains the data from 27 entities monitored by 55 metrics. SWaT~\cite{mathur2016swatwadi} is collected from a scaled-down real-world industrial water treatment plant that produces filtered water. In SWaT, operational data under normal circumstances are collected for 7 days, and operational data with attack scenarios are collected under 4 days. In WADI~\cite{mathur2016swatwadi}, operational data under normal circumstances are collected for 14 days, and operational data with attack scenarios are collected under 2 days. This dataset is collected from the WADI testbed, an extension of the SWaT testbed.

\end{document}